\pdfoutput=1

\documentclass[11pt]{article}

\usepackage{ACL2023}

\usepackage{times}
\usepackage{latexsym}

\usepackage[T1]{fontenc}

\usepackage[utf8]{inputenc}

\usepackage{microtype}

\usepackage{inconsolata}
\usepackage{amsmath}
\usepackage{amsthm}
\usepackage{caption}
\usepackage{subcaption}
\usepackage{bbding}
\usepackage{pifont}
\usepackage{amssymb}
\usepackage{multirow}
\usepackage{graphicx}

\title{Debiasing Stance Detection Models with Counterfactual Reasoning and Adversarial Bias Learning}

\author{Jianhua Yuan \and Yanyan Zhao \and Bing Qin \\
  Research Center for Social Computing and Information Retrieval, \\
    Faculty of Computing, \\
  Harbin Institute of Technology, Harbin  150001, China \\
  \texttt{\{jhyuan, yyzhao, qinb\}@ir.hit.edu.cn} \\}

\begin{document}
\maketitle
\begin{abstract}
Stance detection models may tend to rely on dataset bias in the text part as a shortcut and thus fail to sufficiently learn the interaction between the targets and texts. Recent debiasing methods usually treated features learned by small models or big models at earlier steps as bias features and proposed to exclude the branch learning those bias features during inference. However, most of these methods fail to disentangle the ``good'' stance features and ``bad'' bias features in the text part.  In this paper, we investigate how to mitigate dataset bias in stance detection. Motivated by causal effects, we leverage a novel counterfactual inference framework, which enables us to capture the dataset bias in the text part as the direct causal effect of the text on stances and reduce the dataset bias in the text part by subtracting the direct text effect from the total causal effect. We novelly model bias features as features that correlate with the stance labels but fail on intermediate stance reasoning subtasks and propose an adversarial bias learning module to more accurately model the bias.  To verify whether our model could better model the interaction between texts and targets, we test our model on recently proposed test sets to evaluate the understanding of the task from various aspects. Experiments demonstrate that our proposed method (1) could better model the bias features, and (2) outperforms existing debiasing baselines on both the original dataset and most of the newly constructed test sets. 
\end{abstract}

\section{Introduction}

\begin{figure}
    \centering
    \includegraphics[width=3in]{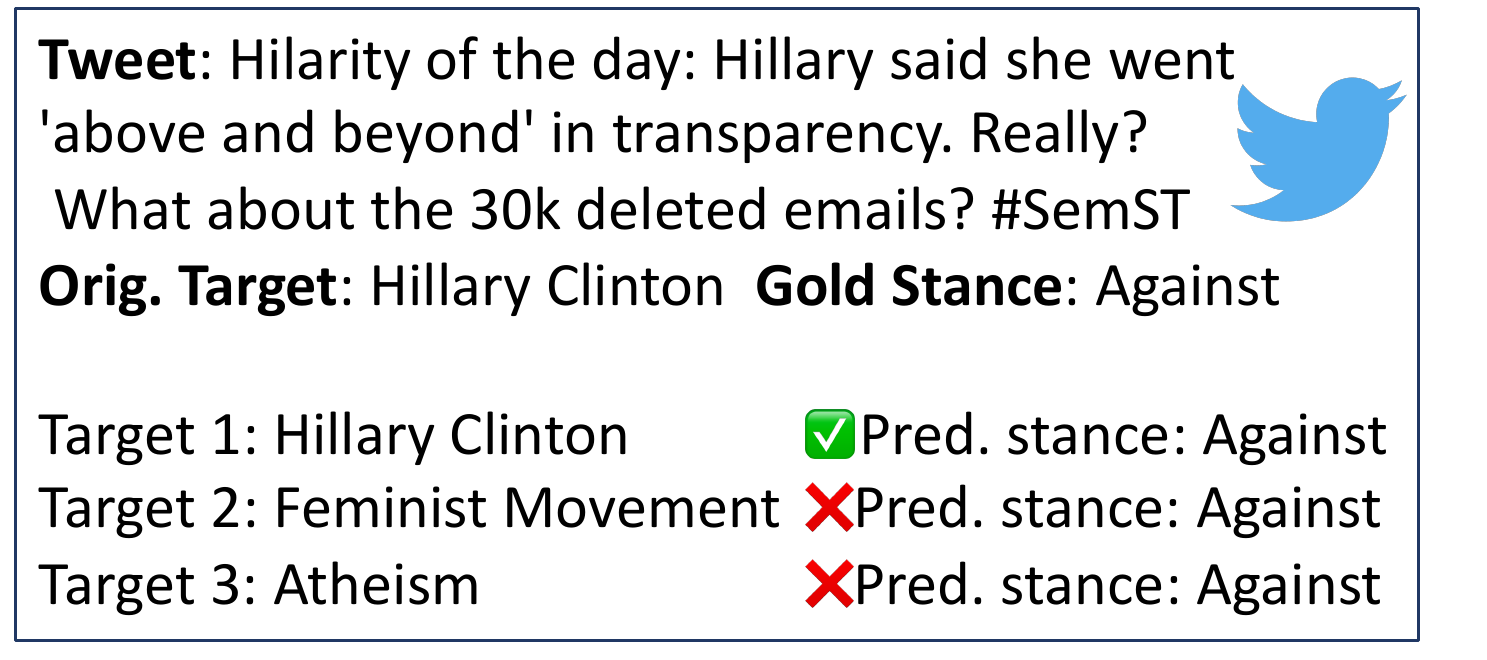}
    \caption{An example illustrates that the BERT model does not change predictions of stance according to the target.}
    \label{fig:bert_pred_unchanged}
\end{figure}

The task of stance detection aims to predict the stance of the text towards the given target. It is crucial for various downstream tasks including fact verification, rumor detection, etc. It has a wide application in political analysis and product reviewing. Existing works usually treat this task as a text pair classification problem and many design target-aware structures \cite{augenstein-etal-2016-stance} to learn target-aware stance representations. However, \cite{kaushal-etal-2021-twt} have shown that those models \cite{du-ijcai2017-557,devlin2019bert} could achieve superior performances even not using the target at all. As shown in Figure  \ref{fig:bert_pred_unchanged}, a BERT model does not change its predictions even if the target is not referred to in the tweet. Existing end-to-end stance detection models treat the stance deliberation process as a black-box  and are prone to rely on bias features in the dataset instead of learning the underlying task. In consequence, those models perform poorly on out-of-distribution datasets \cite{kaushal-etal-2021-twt} and unseen targets, which calls for debiasing stance detection models to get rid of spurious correlations in the datasets.

While \citet{kaushal-etal-2021-twt} made the first attempt to reveal dataset bias in stance detection, mitigating bias in other natural language understanding (NLU) tasks has been extensively explored. One line of work exploits data augmentation-based methods to break the spurious correlations in original distributions, which usually involves additional data collection/selection/annotation. Another line of work treated features learned by small models or big models at earlier steps as bias features and proposed to exclude the branch learning those bias features during inference. However, valuable semantics learned may also be excluded during re-weighting.

To exclude spurious correlation between the text and stance labels while keeping the good context in the text part, we resort to causal inference. In particular, we utilize a causal graph to inspect the relationship between text, target, and stance labels. From the graph, we can discover both direct and indirect effects of the text on stance labels. The former implies a shortcut to the prediction, which exhibits spurious correlations in the text part. The latter, in contrast, extracts reliable interactions between the text and target. 

To disentangle the direct and indirect effects of text and mitigate the bad direct effect, we devise a counterfactual reasoning method. During training, we add an extra branch from text to stance prediction. In the inference, we estimate the direct effect of text by counterfactual reasoning, which imagines \textit{what the prediction would be if the model had only seen the effect of texts}. After that, we mitigate the direct effect and utilize the indirect one of target-text interactions to alleviate the influence of spurious correlations in texts. To better depict the direct effect of text and keep more good context in the text part, we further assume features that are useful for stance prediction while harmful for intermediate tasks of stance detection as bias features. We add an additional adversarial bias learning module to concretize this idea. Our proposed method, CRAB (\underline{c}ounterfactual \underline{r}easoning and \underline{a}dversarial \underline{b}ias learning), achieves superior performances on both original dataset and several adversarial test sets, compared to existing stance detection models and classic debiasing methods.

To summarize, our contributions are three-fold:
\begin{itemize}
    \item We make the first attempt to utilize counterfactual reasoning in stance detection to mitigate spurious correlations between the texts and stance labels.
    \item We propose a novel adversarial bias learning module to more precisely depict the bias features in the text part. 
    \item Experimental results on several benchmark datasets demonstrate the effectiveness of our proposed methods.

\end{itemize}

\section{Related Work}

\subsection{Stance Detection}

Recently, detecting stances in texts from social media platforms have attracted a lot of attention. Compared to traditional sentiment analysis tasks, stance detection is more challenging as the given target may not appear in the text. Inferring the relations between the given target and the opinioned entity usually requires rich world knowledge. In this paper, we focus on the single target stance detection on tweets where each tweet is given one target. Various methods \cite{augenstein-etal-2016-stance,du-ijcai2017-557,popat-etal-2019-stancy} have been proposed to model the inter-dependency between the target and tweet. However, \citet{kaushal-etal-2021-twt} recently noted that current stance detection models relied heavily on bias features in existing stance detection datasets, which makes it necessary to develop stance detection specific debiasing methods to combat these biases. 

In this work, we also study the problem of dataset bias in stance detection and make the first attempt to tackle the problem from a causal view.

\subsection{Debiasing Dataset Bias in NLU}

Recently, much research work had shown that neural models could achieve good performance by simply leveraging dataset bias in existing natural language tasks, e.g. NLI \cite{gururangan2018annotation,poliak-etal-2018-hypothesis}, question answering \cite{mudrakarta-etal-2018-model}, VQA \cite{Agrawal_2018_CVPR}, fact verification \cite{schuster2019towards} and sentiment analysis \cite{wang-culotta-2020-identifying,Wang_Culotta_2021,kaushal-etal-2021-twt,yan-etal-2021-position,yang-etal-2021-exploring}.  To avoid overfitting to bias features, one line of work \citet{tu2020empirical}  proposed to use multi-task learning based-approach by introducing auxiliary tasks like paraphrase identification to avoid overfitting to bias features. Another line of work \cite{clark2019don,mahabadi2020end,utama2020mind,ghaddar-etal-2021-end} employed a weak model to learn bias features and down-weight these features or examples containing those features to achieve debiased training. 

Different from those task-agnostic debiasing methods, we utilize the task knowledge of stance detection and incorporate intermediate stance reasoning sub-tasks to guide the modeling of bias features. We employ weak models to capture features useful for stance detection but not useful for intermediate sub-tasks as bias features through adversarial learning \cite{belinkov-etal-2019-adversarial}.

\subsection{Causally Inspired NLP}
Counterfactual reasoning and causal inference have inspired several recent studies in natural language processing, including visual question answering \cite{niu2020counterfactual}, natural language inference \cite{Tian_Cao_Zhang_Xing_2022}, question answering \cite{wang-etal-2021-counterfactual-adversarial}, named entity recognition \cite{zhang-etal-2021-de}, and text classification \cite{wang-culotta-2020-identifying, qian-etal-2021-counterfactual,Choi_Jeong_Han_Hwang_2022}. Different from existing task-agnostic counterfactual frameworks, our model incorporates an adversarial bias learning module that leverages intermediate sub-tasks of stance detection to learn bias more accurately.

\section{Approach}
In this section, we first introduce the key concepts of counterfactual inference. Then, we describe the causal view of the stance detection process, followed by an introduction to the novel task-aware adversarial bias learning module. Lastly, we detail how we train and make inference.

\begin{figure}
    \centering
    \includegraphics[width=0.9\linewidth]{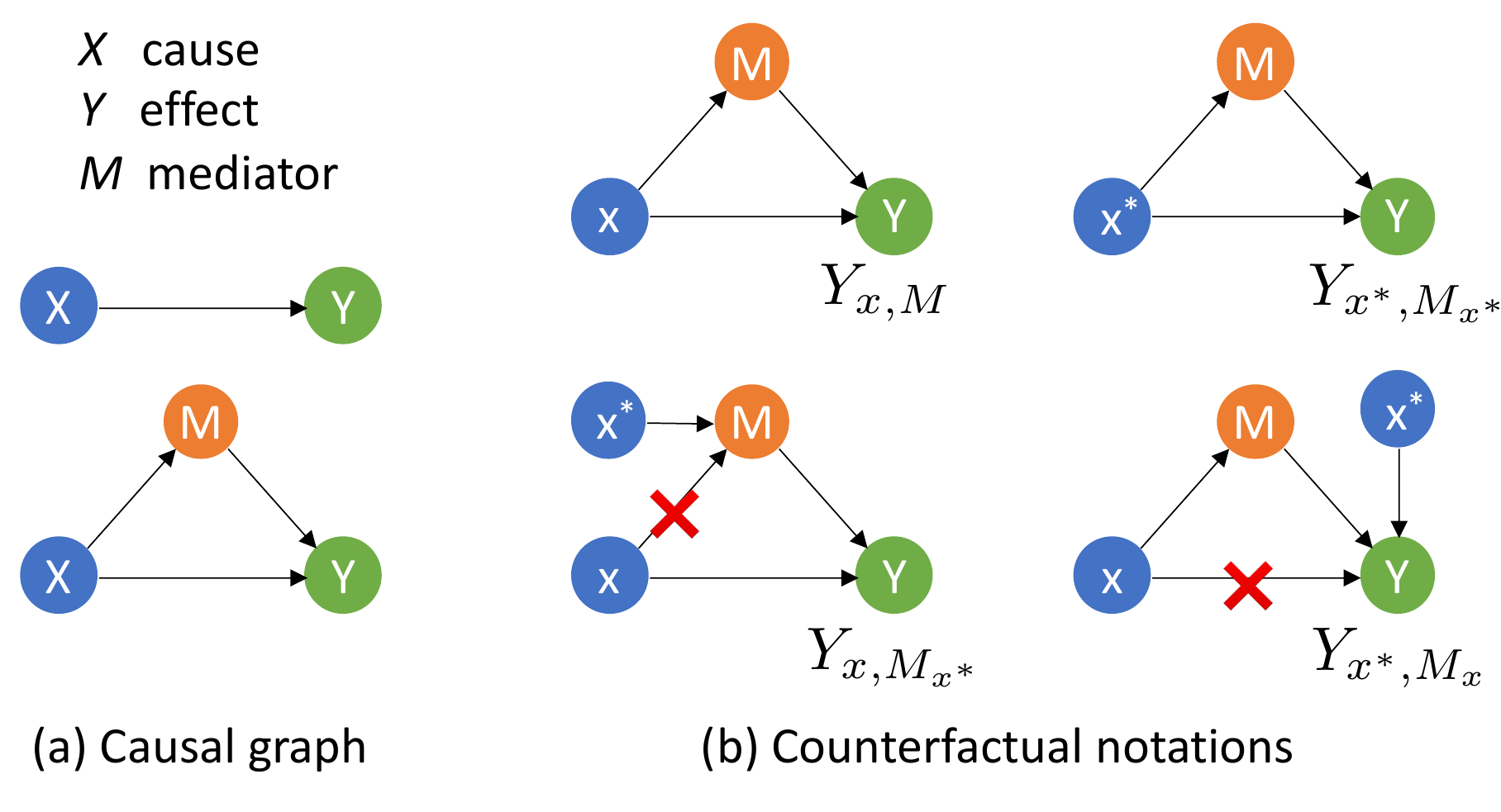}
    \caption{(a) Example of causal graph. (b)Example of counterfactual notations}
    \label{fig:causal_graph_preliminary}    

\end{figure}

\subsection{Preliminary}

\paragraph{Causal graph} is a directed acyclic graph $G={V, E}$, where $V$ denotes the set of variables and $E$ represents the causality among those variables. If the variable $X$ has a \textit{direct} effect on variable $Y$, we say that $X$ is the parent of $Y$, i.e., $X \rightarrow Y$. If $X$ has an \textit{indirect} effect on $Y$ via the variable $M$, we say that $M$ serves as a \textit{mediator} between $X$ and $Y$, i.e., $X \rightarrow M \rightarrow Y$. Figure \ref{fig:causal_graph_preliminary}(a) shows an example of causal graph with 2 and 3 variables respectively.

\paragraph{Counterfactual} means ``counter to the fact'' \cite{roese1997counterfactual}, which assigns non-factual combination of values to variables. Counterfactual notations are used to translate causal graphs into formulas. Take Figure \ref{fig:causal_graph_preliminary}(b) as an example, if $X$ is set to $x$ and $M$ is set to $m$, the value of $Y$ can be denoted as:
\begin{equation}
    Y_{x,m} = Y(X=x, M=m)
\end{equation}
where the value of mediator can be obtained through $m = M_x = M(X=x)$. In the counterfactual scenario, $X$ is simultaneously set to different values for $M$ and $Y$. For example, $Y_{x, M_{x^\ast}}$ refers to the situation where $X$ is set to $x$ and $M$ is set to value where $X$ is $x^\ast$, i.e., 
\begin{equation}
\label{equ:counterfactual}
Y_{x, M_{x^\ast}} = Y(X=x, M=M(X=x^\ast))
\end{equation}

\paragraph{Cause effect} reflects the comparisons between two potential outcomes of the same individual given two different treatments. For example in Figure \ref{fig:causal_graph_preliminary}(b), supposed that $X=x$ means ``under treatment condition'' and $X=x^\ast$ means ``under no treatment condition''. Then by comparing between the factual scenario $X = x$ and counterfactual scenario $X = x^\ast$, we can get the total effect of $X = x$ on $Y$ as:
\begin{equation}
    TE = Y_{x, M_x} - Y_{x^\ast, M_{x^\ast}}
\end{equation}

Total effect can be further decomposed into natural direct effect (NDE) and total indirect effect (TIE), which represents the effect through the direct path $X \rightarrow Y$ and indirect path $X \rightarrow M \rightarrow Y$ respectively. NDE represents the effect of $X$ on $Y$ when the mediator $M$ is blocked. It reflects the increase in the outcome $Y$ when $X$ is changing from $x^\ast$ to $x$ while $M$ is set to the value when $X=x^\ast$:
\begin{equation}
    NDE = Y_{x, M_{x^\ast}} - Y_{x^\ast, M_{x^\ast}}
\end{equation}
where $Y_{x, M_{x^\ast}} = Y(X=x, M=M(x^\ast))$. The calculation of $Y_{x, M_{x^\ast}}$ is a counterfactual inference since  variable $X$ is set to different values for different paths. 

TIE is the difference between TE and NDE, defined as:
$$
TIE = TE - NDE = Y_{x, M_x} - Y_{x, M_{x^\ast}}
$$
which denotes the effect of $X$ on $Y$ through indirect path $X \rightarrow M \rightarrow Y$.

Next, we will discuss the meanings of these effects in stance detection.

\subsection{Causal Look at Stance Detection}
\begin{figure}
    \centering
    \includegraphics[width=\linewidth]{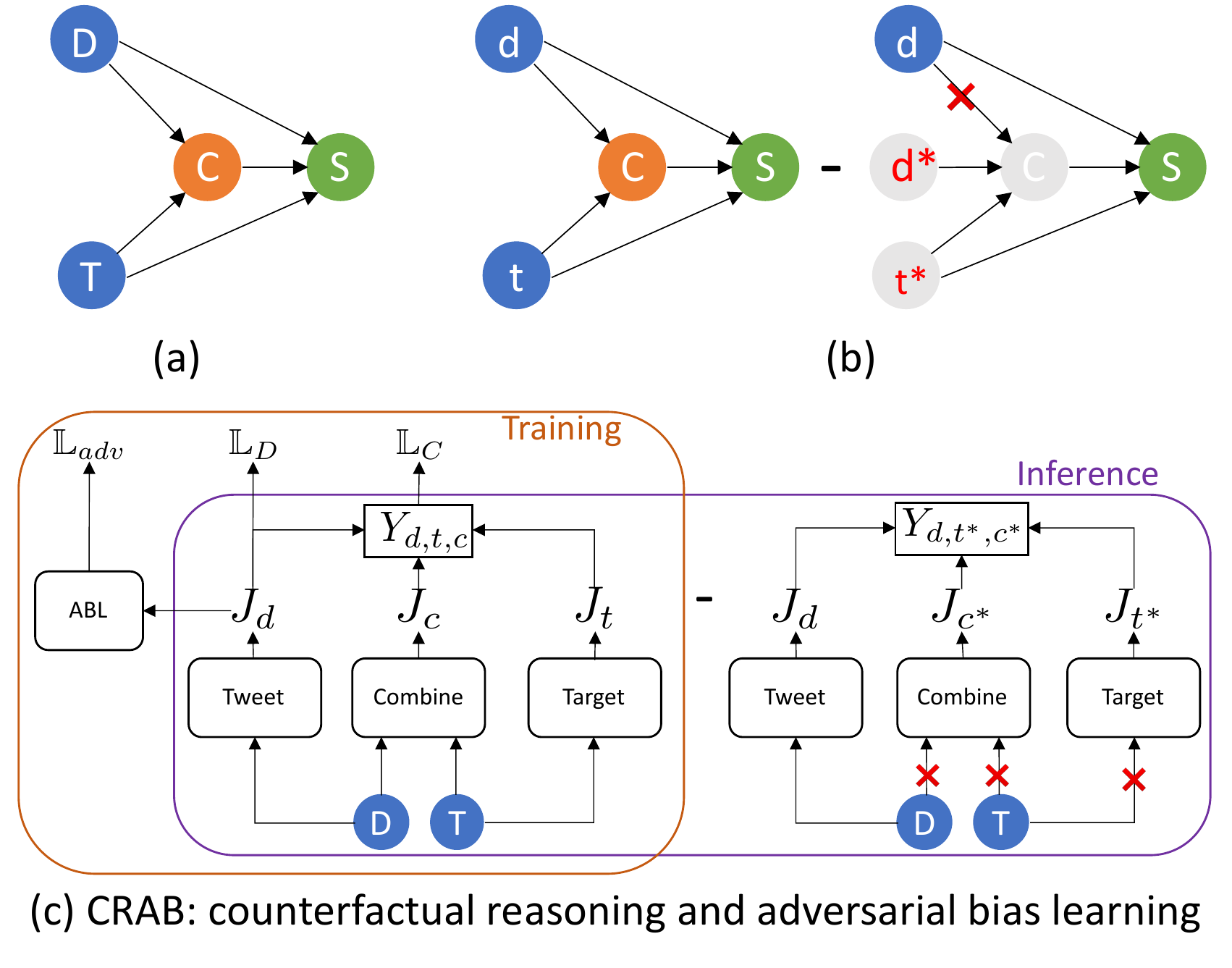}
    \caption{(a) Causal graph for stance detection. D: text or tweet. T: target. C: target-tweet interaction knowledge. S: stance. (b) Comparison between conventional stance detection (left) and counterfactual stance detection (right). (c) the proposed CRAB model.}
    \label{fig:full_model_all}    
    
\end{figure}

Following previous work, we formulate the stance detection task as a three-way classification problem. Given a text $D=d$ and a target $T=t$, stance detection models are required to determine the stance of the text $d$ towards the given target $t$.

The causal graph of stance detection is illustrate in Figure \ref{fig:full_model_all}(a). The effect of text $D$ and target $T$ on stance $S$ can be divided into text-only impact and text-target-fusion impact. The text-only impact captures the direct effect of tweet $D$ on stance $S$ via $D \rightarrow S$. The text-target-fusion impact captures the indirect impact of $D$ and $T$ on $S$ via the target-text combined module $C$, i.e., $D,T \rightarrow C \rightarrow S$. We aim to mitigate the dataset set bias in the tweet part by excluding the pure tweet effect on $D \rightarrow S$.

Following the notations in Eq.\ref{equ:counterfactual}, we define the probability of the stance $s$ would obtain if the text $D$ is set $d$ and the target $T$ is set to $t$ as :
\begin{equation}
\label{equ:cause_effect_stance}
Y_{d,t,c}(s\footnote{We omit $s$ for simplicity without the loss of generality.}) = Y(D=d, T=t, C=c)
\end{equation}
where the notation of c is denoted as $c=C_{d,t} = C(D=d,T=t)$.
Then the total effect (TE) of the tweet $D=d$ and target $T=t$ on stance $s$ can be written as:
\begin{equation}
\label{equ:te_stance}
TE = Y_{d,t,c} - Y_{d^\ast,t^\ast,c^\ast}
\end{equation}
where $d^\ast$ and $t^\ast$ represent the no-treatment condition where $d$ and $t$ are set to void values and $c^\ast=C(d^\ast,t^\ast)$.

As mentioned earlier, there exists dataset bias in the tweet part which allows stance detection models to achieve superior performance without using the target. Such dataset bias hinders stance detection models to perform effective stance reasoning between targets and texts. Thus, we expect to exclude the direct impact of texts from stance detection models. To this end, we utilize counterfactual reasoning to estimate the cause effect of pure tweet $D=d$ on stance $s$ by blocking the effect of $C$ and $T$. We use counterfactual stance detection to describe scenarios where $D$ is set to $d$ and $C=c^\ast$ when $D$ is set to $d^\ast$ and $T$ is set to $t^\ast$. As the impact of mediator $C$ to input is blocked, the model makes stance predictions solely based on the tweet part.

Thus, we obtain the natural direct effect (NDE) of $D$ on $S$ by comparing the counterfactual stance detection to the no-treatment condition as:
\begin{equation}
\label{equ:nde_stance}
NDE = Y_{d,t^\ast,c^\ast} - Y_{d^\ast,t^\ast,c^\ast}
\end{equation}
Since the effects of $T$ and $C$ on the stance are blocked, NDE could explicitly capture the dataset bias in the tweet part. We further reduce the text-only bias by subtracting NDE from TE as follows in Figure \ref{fig:full_model_all}(b):
\begin{equation}
\label{equ:tie_stance}
TIE = TE -NDE =  Y_{d,t,c} - Y_{d,t^\ast,c^\ast}
\end{equation}
We then select the stance with maximum TIE for inference, which differs from traditional factual stance detection that is based on the posterior probability i.e., $P(s|d,t)$.

\subsection{Adversarial Bias Learning}
\begin{figure}
    \centering
    \includegraphics[width=0.6\linewidth]{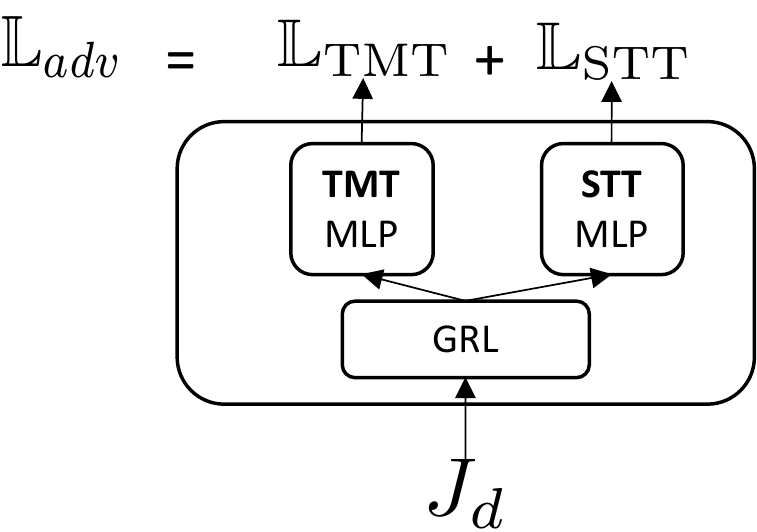}
    \caption{Adversarial bias learning module}
    \label{fig:adv_module}    

\end{figure}

To preserve more context information while removing dataset bias in the text part, we further propose an adversarial bias learning module to capture the bias more accurately. Previous task-agnostic debiasing methods \cite{clark2019don,mahabadi2020end,du-etal-2021-towards} usually treat features learned by small models or features learned by large models at earlier steps as bias features. In contrast, from the perspective of stance detection, we regard features that are useful for the final stance prediction while not valid for intermediate stance reasoning sub-tasks as bias features. In the following, we detail how we choose sub-tasks and perform adversarial bias learning.

\subsubsection{Intermediate Sub-tasks}
Given a pair <target, text> as input, a common stance detection process usually involves 1) identifying the target mentioned in the text, 2) finding corresponding opinion expressions towards the mentioned target and deciding their sentiments, 3) deciding the relationship between the mentioned target and the given target, and 4) making final stance stance predictions. However, opinion spans and target spans are hard and expensive to annotate in those informal texts on social media. As a result, it is impractical to directly apply the above tasks for more accurate bias learning. 

To this end, we seek to simplify those intermediate sub-tasks into easier ones that only require sentence-level annotations instead of token-level ones. Specifically, we adopt two binary classification sub-tasks, which are 1) \textbf{TMT}, whether the given \underline{t}arget is \underline{m}entioned in the \underline{t}ext and 2) \textbf{STT}, whether the text express any \underline{s}tance \underline{t}owards the given \underline{t}arget. TMT is designed to make stance detection models aware of the (both explicit and implicit) existence of the target in the text. STT requires stance detection models to differ the \textit{None} stance from other stances. One advantage of adopting TMT and STT as intermediate sub-tasks is that they can not be solely solved using the text part. The other advantage is that labels for TMT and STT can be automatically inferred using simple text matching and existing stance labels (see Appendix for details on the acquisition of those annotations).

\subsubsection{Adversarial Learning}
 
In the causal graph of stance detection, the $D \rightarrow S$ branch could capture the dataset bias in the text $d$. Thus, we perform adversarial bias learning on this branch in the hope to capture bias features useful for finding stance detection while not useful for other intermediate sub-tasks. Motivated by the recent success of adversarial training in sentiment analysis and debiasing NLU models, we use representations of the input text to predict the stance labels as well as intermediate sub-tasks. We feed $h_d$, the representation of the input text, into an MLP classifier for stance prediction. As shown in Fig. \ref{fig:adv_module}, we pass $h_d$ through a gradient reversal layer.
\begin{eqnarray}
    \label{equ:grl}
    h_d &=& Encoder(d) \\
    h_{grl} &=& GRL(h_d)
\end{eqnarray}
Then we feed $h_{grl}$ into another two MLP classifiers for TMT and STT respectively. During the forward propagation, GRL acts as an identity transform, making no changes to $h_d$. During the back-propagation pass, GRL takes the gradient from the subsequent level, reverses the gradient, and passes it to the preceding layer. The gradients from TMT and STT classiﬁcation will not be correctly sent back to the encoder part, thus making it hard to learn features useful for TMT and STT. In this way, the reversed gradients from TMT and STT classifiers will drive $h_d$ to contain features less useful for intermediate stance reasoning tasks. And the representations from $D \rightarrow S$ become maximally informative for stance detection while simultaneously minimizing the ability of $h_d$ to accurately perform intermediate reasoning tasks. The adversarial bias learning module is optimized by minimizing the cross-entropy losses $\mathbb{L}_{TMT}$ and $\mathbb{L}_{STT}$  over TMT and STT tasks jointly.

\subsection{Training and Inference}
To sum, we propose CRAB, a model with \underline{c}ounterfactual \underline{r}easoning and \underline{a}dversarial \underline{b}ias learning, to mitigate dataset bias in stance detection. We formulate each branch of Figure \ref{fig:full_model_all}(a) as a neural model. The calculation of stance score $Y_{d,t,c}$ is parameterized as follows:
\begin{equation}
\label{equ:para_stance_score}
Y_{d,t,c} =  F(J_d,J_t,J_c)
\end{equation}
where $J_d = F_D(d)$ is the text-only branch (i.e., $D \rightarrow S$), $J_t = F_T(t)$ is the target-only branch (i.e., $D \rightarrow S$), $J_c = F_C(c)$ is the combined feature branch (i.e., $C \rightarrow S$). $F$ is the fusion function to obtain the final stance scores. 

As neural models cannot deal with no-treatment conditions where the inputs are void, we assume that the model will make random guess with equal probability under no-treatment conditions. Therefore, we represent no-treatment conditions with the same constant $a$ which is a learnable parameter. This intuition makes senses as humans would like to make a random guess when we have no information about the stance detection task. We fuse the representation as follows:
\begin{equation}
\label{equ:para_stance}
F(d,t,c) =  \log(\sigma(J_d + J_t +J_c))
\end{equation}

\paragraph{Training.} The training process is illustrated in Figure \ref{fig:full_model_all}(c). Specifically, our model is optimized by minimizing the cross-entropy losses over the scores $J_c$ and $J_d$.
\begin{equation}
\label{equ:loss}
\mathbb{L}_{cls} = \mathbb{L}_{C} + \mathbb{L}_{D} + \lambda_1\mathbb{L}_{TMT} + \lambda_2\mathbb{L}_{STT}
\end{equation}
where $\mathbb{L}_{C}$ and $\mathbb{L}_{D}$ are losses over $J_c$ and $J_d$ respectively, $\lambda_1$ and $\lambda_2$ are coefficients controlling the impact of two sub-tasks respectively.

As text-only models perform similarly to a model that utilizes both the target and text, we assume that the sharpness of text-only branch should be similar to that of the text-target-combined branch. Thus, we use Kullback-Leibler divergence to estimate $c$:
\begin{equation}
\label{equ:loss_kl}
\mathbb{L}_{kl} = \frac{1}{\mathbb{A}}\sum_{a \in \mathbb{A}}-p(a|d,t,c) \log p(a|d,t^\ast,c^\ast)
\end{equation}
where $p(a|d,t,c)=softmax(J_{d,t,c})$ and $p(a|d,t^\ast,c^\ast)=softmax(J_{d,t^\ast,c^\ast})$. Only $a$ is updated when minimizing $\mathbb{L}_{kl}$. The final loss is the combination of $\mathbb{L}_{cls}$ and $\mathbb{L}_{kl}$.

\paragraph{Inference.} In the test stage, we use the debiased  effect for inference, which is implemented as:
\begin{equation}
\label{equ:inference}
\begin{split}
TIE &= TE - NDE \\ 
&= Y_{d,t,c} - Y_{d,t^\ast,c^\ast} \\
&=F(J_d, J_t, J_c)-F(J_d, J_{t^\ast}, J_{c^\ast})
\end{split}
\end{equation}

\section{Experimental Setups}

\subsection{Datasets}
To thoroughly assess whether a model understands the stance detection task or just simply over-fitting to bias features, we test on our model on both original semeval2016 test data and newly proposed hard test data by \citep{yuan-etal-2022-ssr}.

The detailed statistics of this dataset are shown in Table \ref{tab:orig_dataset}.

\begin{table} 
\small
\centering
\begin{tabular}{cccc}
\hline
Target  & \#total & \#Train  & \#Test  \\ 
\hline
Atheism & 733 & 513 & 220 \\
Climate Change & 564 & 395 & 169 \\
Feminist Movement & 949 & 664 & 285  \\
Hillary Clinton & 984 & 689 & 295\\
Abortion  & 933 & 653& 280\\
\hline
All  & 4163 & 2914& 1249 \\ 
\hline
\end{tabular}
\caption{Statistics of SemEval2016 Task 6 Subtask A.} 
\label{tab:orig_dataset}
\end{table}

\begin{table}
    \small
	\centering
	\begin{tabular}{ l | c}
		\hline
		New Test Sets & Number \\
		\hline
		Tweet\_only Failed (TOF) & 319  \\
		PMI-tail (PMI) & 403 \\
		Opinion Towards (OT) & 425 \\
		\hline 
		Donald Trump (DT) & 707 \\
		\hline
		Target Replaced (Replaced) & 3978 \\
		Target Negated (Negated) & 1249 \\
		\hline
	\end{tabular}
	\caption{Statistics of the used hard test data by \cite{yuan-etal-2022-ssr}.}
	\label{tab:hard_set}
\end{table}

\subsection{Evaluation Metrics}
We adopt the macro-average of F1-score across targets as the evaluation metric and we report the averaged results of 5 random seeds for all experiments. Similar to previous work, we adopt the macro-average of F1-score across targets as the evaluation metric, which is calculated as:
\begin{eqnarray}
    F_{Favor} &=& \frac{2 P_{Favor} R_{Favor}}{P_{Favor} + R_{Favor}} \\
    F_{Against} &=& \frac{2 P_{Against} R_{Against}}{P_{Against} + R_{Against}} \\
    F_{macro} &=& \frac{2 (F_{Against} + F_{Favor})}{2}
\end{eqnarray}
where $P$ and $R$ are precision and recall respectively. Then the average of $F_{Favor}$ and $F_{Against}$ is calculated as the final metrics $F_{macro}$. We report the averaged results of 5 random seeds for all experiments.

\subsection{Implementation}
We use the official train/test split. We randomly select 15\% of samples from the training data for validation. We adopt the uncased version of BERT$_\text{base}$ for all our experiments. We fine-tune BERT$_\text{base}$ model with Adam optimizer. The dropout rate is set to 0.5 for all parameters. The learning rate is chosen from $\lbrace 1, 2, 3, 4, 5 \rbrace \times 10^{-5}$ and the batch size for training is set to 8. We choose $\lambda_1$ and $\lambda_2$ from [0.1, 1.0] with a step size of 0.1. The final choices of all hyper-parameters are selected according to performance on the validation set. $\lambda_1$ and $\lambda_2$ are set to 0.1 and 0.2 respectively. The learning rate is set to $5\times 10^{-5}$.

\subsection{Baselines}
\paragraph{Stance Detection Models}
Methods on Stance detection: 1) BERT$_\text{wt}$ and BERT$_\text{nt}$, which are based on BERT$_\text{base}$. BERT$_\text{nt}$ only uses the tweet as input while BERT$_\text{wt}$ takes the <target, tweet> pair as inputs. 2) TAN \cite{du-ijcai2017-557}, which is an LSTM based model that incorporates target-specific attention. Following \citet{kaushal-etal-2021-twt}, we adopt the BERT version of TAN \cite{kaushal-etal-2021-twt}. 3) Stancy \cite{popat-etal-2019-stancy}, which is a BERT-based model with an additional cosine similarity score between the tweet representation and the target-tweet pair representation.

\paragraph{Debiasing Models}
Apart from sophisticated stance detection models, we also compare with recent debiasing methods for natural language inference and fact verification tasks. These methods are 1) Product-of-expert (PoE) \cite{clark2019don}, which combines the learned probabilities of a bias-only model and full model using PoE. 2) LMH \cite{clark2019don}, which explicitly determines how much to trust the bias in PoE and employs an entropy-based regularization to encourage the bias component to be non-uniform. 3) E2E PoE \cite{mahabadi2020end}, which proposes an end-to-end training version of PoE. 4) Conf-Reg \cite{utama2020mind}, which utilizes signals from bias models to scale the confidence of models' predictions. 5) SSR \cite{yuan-etal-2022-ssr}, which also incorporates simplified stance reasoning process through multi-task learning. In contrast, we use these subtasks in an adversarial way.

\section{Results and Analysis}
\begin{table}
\small
\centering
\begin{tabular}{l | c | c c c }
\hline
Models     & Original & TOF & PMI & OT        \\
\hline

BERT$_{nt}$ & 67.84 & 73.9 & 50.8 & 38.15 \\ 
BERT$_{wt}$ & 69.21 & 85.56 & 51.66 & 42.62\\ 
TAN & 68.44 & 86.51 & 59.31 & 43.88\\ 
Stancy & 70.3 &  95.34 & 59.78 & 44.67 \\
\hline
PoE & 68.96 & 94.06 & 55.15 & 41.53 \\
LMH & 64.88 & 82.69 & 47.46 & 33.81 \\	
E2E-PoE & 61.68 & 84.88 & 53.93 & 39.08  \\

Conf-Reg & 70.24 & 90.45 & \underline{61.26} & 41.16   \\
\hline
SSR  & \underline{71.36} & \textbf{96.47} & 56.08 & \underline{46.58} \\
\hline
CRAB (ours) & \textbf{71.45} & \underline{94.47} & \textbf{64.41} & \textbf{46.62} \\
\hline
\end{tabular}
\caption{Results on the SemEval test dataset and three subsets. The average of $F_{Favor}$ and $F_{Against}$ is adopted as the evaluation metric. For comparison with other stance detection models on each target, please refer to Table \ref{tab:orig_dataset_full} in Appendix for details.}
\label{tab:main_result}
\end{table}

\subsection{Main Results}

\subsubsection{On Original Test Set and its Subsets}

As illustrated in Table \ref{tab:main_result}, the BERT$_{nt}$ model, which neglects target information, achieves similar performance to the BERT$_{wt}$ model that considers target on the original test set, proving the existence of dataset bias in the tweet part. The performances of stance detection methods are relatively similar while debiasing methods like LMH and E2E-PoE suffer from a significant performance drop. In contrast, our model outperforms both debiasing baselines and stance detection baselines on the original test set.

Though the TOF test set is constructed by selecting examples where three BERT models with different seeds fail, another BERT model with a new seed could reach the performance of 73.9\%. Our model outperforms all debiasing methods and most of the stance detection baselines, showing that our model could better deliberate stance where target information is crucial.

On the PMI set where long-tail features exist, the BERT$_{nt}$ model and BERT$_{wt}$ achieve similar results, which illustrates that BERT$_{wt}$ may not utilize target-tweet interaction well and instead mostly exploits bias features in the tweet part. Our model achieves the best performances and outperforms the BERT$_{wt}$ model by over 12\%, demonstrating a significant reduction in reliance on the dataset bias features in the tweet part.

Since both our model and Stancy have explicit task-aware modeling of the target, they outperform other methods that do not utilize such task knowledge on the OT set. The targets in the OT set are usually implicitly mentioned in the tweet and it is much harder than the explicitly mentioned counterpart according to the performances of stance detection models. Our model is better at deliberating the stance of both explicitly and implicitly mentioned targets.

\subsubsection{On New Test Sets}

\begin{table}
    \small
    \centering
    \begin{tabular}{l|c|c|c c}
        \hline
        Model & DT & Replaced & Negated  \\
        \hline
        BERT$_{nt}$ & 11.42 & 32 & 17.59 \\
        BERT$_{wt}$ & 28.12 & 47.08 & 17.38 \\
        TAN & 27.09 & 35.55 & \underline{20.05} \\
        Stancy & 32.34 & 49.5 & 19.67 \\
        \hline
        PoE  & 19.42 & 46.7 & 19.54 \\
        LMH  & 36.76 & 34.97 & \textbf{25.36} \\	
        E2E-PoE & 33.72 & 33.01 & 19.98 \\
        Conf-Reg & 36.27 & 34.17 & 19.51 \\
        \hline
        SSR & \underline{37.66} & \textbf{59.7} & 17.6 \\
        \hline
        CRAB (ours) & \textbf{40.80} & \underline{56.92 } & 17.99 \\

        \hline
    \end{tabular}
    \caption{Performances of different models on \textit{DT}, \textit{Replaced}, \textit{Negated} test sets. Note that, results on \textit{DT} are not directly comparable to those reported in \cite{allaway-etal-2021-adversarial,10.1145/3485447.3511994} as they used 4,163 pairs for training while we only use 2,914 pairs.}
    \label{tab:hard_result}
\end{table}

As discussed earlier, our model shows superior performance on the original test set where the data distribution is similar to the training one. To test whether stance detection models understand the task instead of simply over-fitting to bias features, we perform experiments on the DT set where the target is unseen during training.

From Table \ref{tab:hard_result}, we can see that in general debiasing methods outperform stance detection models (the only exception is the PoE model). Note that BERT$_{nt}$ performs the worst, which shows that bias features in the training set do not hold on this test set, and overly exploiting bias features indeed hinders the generalization of stance detection models. On the contrary, our model utilizes the task knowledge to more accurately learn bias features and excludes them through counterfactual reasoning, outperforming both stance detection and debiasing baselines by over 4\%.

Similar to the OT set, the adversarially constructed Replaced set also requires a stance detection model to decide whether the target is mentioned in the tweet part. Thus, our model and Stancy again outperform all the other methods. The gap between the BERT$_{nt}$ model and BERT$_{wt}$ model is larger on this test set, showing stronger demands on utilizing target information. By comparing results on the OT and Replaced sets, we can see that debiasing methods do not show any superiority in capturing whether the target is mentioned in the tweet, which implies that task-agnostic debiasing is sub-optimal for robust stance reasoning.

While negation is a crucial component in expressing sentiment and stance, negation in targets or topics is less studied in existing stance detection works. Here, we present preliminary results on the Negated set where the targets are in the form of negations. From Table \ref{tab:hard_result}, we can see that all stance detection models and debiasing methods fail on this task ($<$30\%), implying that the lack of modeling of negations in targets hinders complex stance reasoning. Though augmenting the original training set with examples containing negations \cite{kaushal-etal-2021-twt} can relieve this phenomenon, it may be still far from truly robust stance reasoning through explainable semantic compositions.

\subsection{Ablation Study}
\begin{table}
	\centering
	\small
\begin{tabular}{c c c c c c c}
\hline
\multicolumn{3}{c}{Module} & \multirow{2}{*}{Orig.} & \multirow{2}{*}{TOF} & \multirow{2}{*}{PMI} & \multirow{2}{*}{OT}\\
GRL & STT & TMT &  &  &  & \\
\hline
\ding{52} & \ding{52} & \ding{52} & \textbf{71.45} & \underline{94.47} & \underline{64.41} & 46.62 \\
\ding{56} & \ding{56} & \ding{56} & \underline{70.96} & 93.86 & \textbf{65.74} & \underline{48.13} \\
\ding{56} & \ding{52} & \ding{52} & 67.73 & 92.49 & 62.12 & 45.43 \\
\ding{52} & \ding{56} & \ding{52} & 69.49 & \textbf{94.80} & 62.75 & 46.96 \\
\ding{52} & \ding{52} & \ding{56} & 70.52 & 93.83 & 62.69 & \textbf{48.40} \\

\hline
 &  &  & & DT & Rep. & Neg. \\
\hline
\ding{52} & \ding{52} & \ding{52} & & \textbf{40.80} & \textbf{56.92} &   17.99 \\
\ding{56} & \ding{56} & \ding{56} & & 37.19 & 54.57 & 18.57 \\
\ding{56} & \ding{52} & \ding{52} & & \underline{39.69} & 52.59 & \textbf{20.85} \\
\ding{52} & \ding{56} & \ding{52} & & 34.90 & \underline{55.26} & \underline{19.88} \\
\ding{52} & \ding{52} & \ding{56} &  & 35.50 & 53.93 &  19.11 \\

\hline
\end{tabular}
\caption{Ablation study on the proposed CRAB model. GRL refers to gradient reversal layer, STT and TMT are two intermediate task of stance detection.}
\label{tab:ablation}
\end{table}

To explore the contribution of different components of the proposed CRAB model, we conduct an ablation study on adversarial learning module. By comparing Tabel \ref{tab:main_result}, \ref{tab:hard_result}, and \ref{tab:ablation}, we can see that the CRAB model outperforms the BERT$_{wt}$ model by 1.7\% on the original test set and by 9\% on the DT set. This demonstrates that the counterfactual reasoning framework works to reduce reliance on dataset bias in the training set, improving performance on both in-distribution and out-of-distribution data. By adding the adversarial bias learning module, the performance increases from 70.96\% on original test set and from 37.19\% to 40.80\% on the DT set.

In addition, when we remove the gradient adversarial layer and keep two sub-tasks, the $D \rightarrow S$ branch may learn useful features for both stance prediction and its intermediate sub-tasks, thus excluding those features results in a significant performance drop on the original test set.

We also find that the contributions of two sub-tasks are similar to each other on most test sets. While the TMT sub-task focus on the textual matching of targets and the STT sub-task focus on the semantic matching of targets, the TMT sub-task contributes more on the Replaced set where the target is replaced with another target and the STT sub-task is more useful on the OT set where the target is implicitly mentioned.

\subsection{Case Study and Discussion}
\begin{figure}
	\begin{subfigure}{0.5\textwidth}
		\centering
		\includegraphics[width=0.9\linewidth]{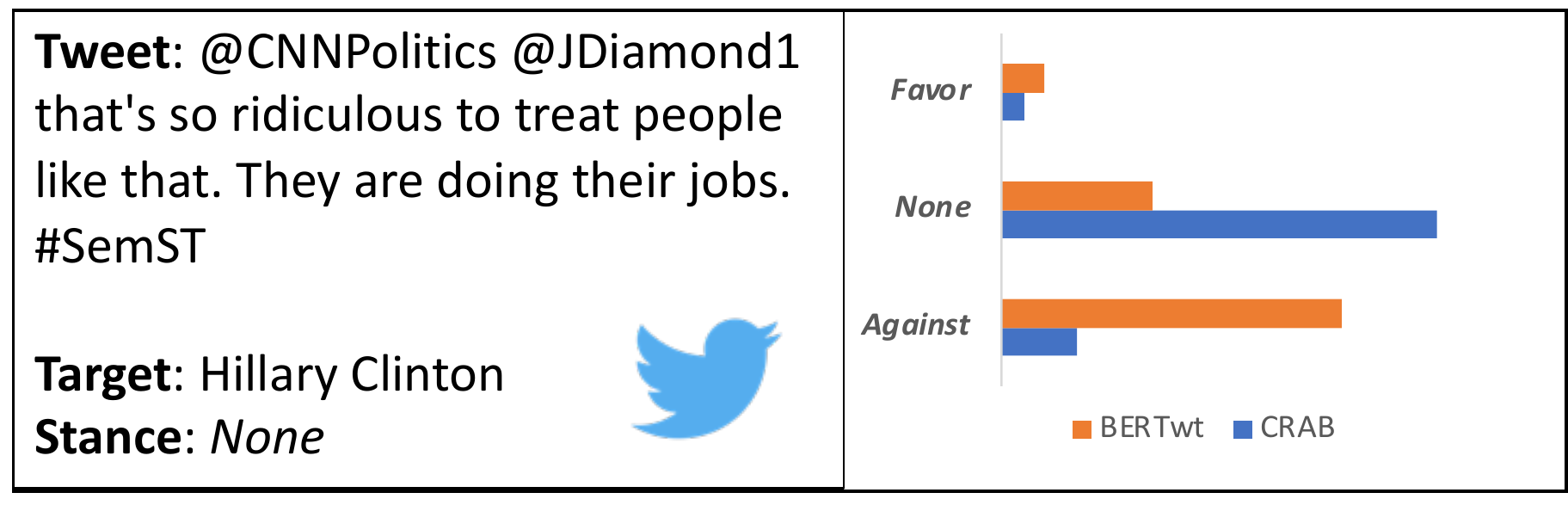}
		\caption{Example 1}
		\label{fig:cm_original}
	\end{subfigure}
	\begin{subfigure}{0.5\textwidth}
		\centering
		\includegraphics[width=0.9\linewidth]{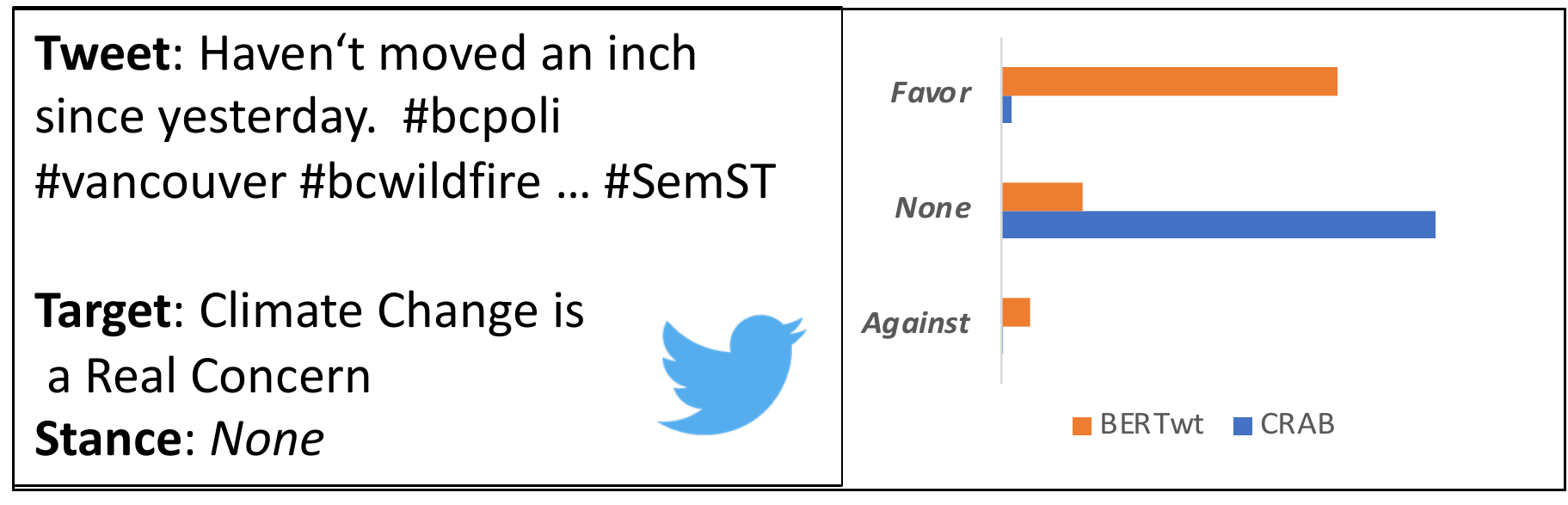}
		\caption{Example 2}
		\label{fig:cm_masked}
	\end{subfigure}
	
	\caption{(Normalized) Stance predictions of the BERT$_{wt}$ model and the proposed CRAB model on two examples. }
	\label{fig:case_study}
\end{figure}

\paragraph{Case Study.} As we can see from Figure \ref{fig:case_study}, in both cases, there are stance-related words like \textit{ridiculous} and \textit{haven't} in the tweet part, the BERT$_{wt}$ model may make false predictions by utilizing those shortcuts. On the contrary, our model makes predictions using the total indirect effect (TIE) where the influence of stance-related words can be excluded and it successfully deliberates the correct stance with high probabilities even with the existence of those misleading words.

\paragraph{Bias in Target.} In the paper, we mainly focus on removing the natural direct effect of $D \rightarrow S$ branch from the final predictions. We also find that the $T \rightarrow S$ branch also contains spurious correlations that may hinder robust stance detection. Using target-wise majority vote could reach the performance of 67\% on the original test set, comparable to the BERT$_{wt}$ line. We leave it as future work to exclude learning of bias features from both $D \rightarrow S$ branch and $T \rightarrow S$ branch.

\section{Conclusion}
In this paper, we make the first attempt to utilize counterfactual reasoning to mitigate dataset bias in stance detection. To more accurately depict the bias features, we propose an adversarial bias learning module that incorporates adversarial learning and two simplified intermediate tasks of stance stance detection to learn features only useful for stance predictions but not useful for two sub-tasks. Experiments on the benchmark stance detection dataset and 6 newly constructed test sets demonstrate the effectiveness of the proposed approach. In the future, we would like to explore excluding bias features from both $D \rightarrow S$ branch and $T \rightarrow S$ branch.

\bibliography{stance}
\bibliographystyle{acl_natbib}

\appendix
\section{Appendix}
\label{sec:appendix}

\begin{table*}[t]
\centering
\small
\begin{tabular}{c|c|cccc|cccc}
\hline
                  &         & \multicolumn{4}{c|}{\% of instances in Train} & \multicolumn{4}{c}{\% of instances in Test} \\
Target            & \#total & \#Train     & Favor    & Against    & None    & \#Test    & Favor    & Against    & None    \\ \hline
Atheism           & 733     & 513         & 17.9     & 59.3       & 22.8    & 220       & 14.5     & 72.7       & 12.7    \\
Climate Change    & 564     & 395         & 53.7     & 3.8        & 42.5    & 169       & 72.8     & 6.5        & 20.7    \\
Feminist Movement & 949     & 664         & 31.6     & 49.4       & 19.0    & 285       & 20.4     & 64.2       & 15.4    \\
Hillary Clinton   & 984     & 689         & 17.1     & 57.0       & 25.8    & 295       & 15.3     & 58.3       & 26.4    \\
Abortion          & 933     & 653         & 18.5     & 54.4       & 27.1    & 280       & 16.4     & 67.5       & 16.1    \\ \hline
All               & 4163    & 2914        & 25.8     & 47.9       & 26.3    & 1249      & 24.3     & 57.3       & 18.4    \\ \hline
\end{tabular}
\caption{Statistics of SemEval2016 Task 6 SubtaskA.} \label{tab:orig_dataset_full}

\end{table*}

\begin{table*}
\centering
\small
\begin{tabular}{lcccccc}
\hline
Models     & AT & CC & FM & HC & LA & Overall        \\
\hline
AT-JSS-Lex \cite{li-caragea-2019-multi} & 69.22 & \underline{59.18} & \underline{61.49} & \underline{68.33} & \textbf{68.41} & \textbf{72.33} \\
CKEMN  \cite{CKEMN} & 62.69   & 53.52  & 61.25   & 64.19   & 64.19    & 69.74 \\ 
\hline
MT-DNN$_{SDL}$ \cite{DBLP:journals/corr/abs-2001-01565} & - & - & - & - & - & 70.18 \\
MT-DNN$_{MDL}$ \cite{DBLP:journals/corr/abs-2001-01565} & - & - & - & - & - & 71.81 \\
MoLE \cite{mole} & - & - & - & - & - & \underline{72.08} \\
ASDA \cite{li-caragea-2021-target} & \textbf{74.93} & - & 56.43 & 67.01 & 61.60 & - \\
MeLT \cite{DBLP:journals/corr/abs-2109-08113} & 66 & \textbf{71} & \textbf{63} & 67 & \underline{66} & - \\
\hline
TAN & \underline{69.72} & 44.32 & 53.26 & 55.79 & 62.75 & 68.44 \\ 
Stancy & 66.08 & 54.67 & 59.91 & 62.0 & 58.69 & 70.3 \\
BERT$_\text{nt}$ & 63.96 & 48.88 & 53.97 & 60.59 & 60.24 & 67.84 \\ 
BERT$_\text{wt}$ & 65.36 & 44.91 & 48.25 & 65.09 & 54.33 & 69.21 \\ 
\hline
CRAB & 66.23 & 54.20 & 58.06 & \textbf{71.00} & 64.87 & 71.45 \\
\hline
\end{tabular}
\caption{Results on the SemEval dataset. The macro average of $F_{Favor}$ and $F_{Against}$ is adopted as the evaluation metric. The best scores for each column are bold and the second best scores are underlined. Results not reported in original papers are marked with -.}
\label{tab:main_result_full}
\end{table*}

\end{document}